\documentclass[runningheads]{llncs}

\usepackage[final, year=2026,ID=3995]{eccv}

\usepackage{eccvabbrv}

\usepackage{graphicx}
\usepackage{booktabs}

\usepackage[accsupp]{axessibility}  

\PassOptionsToPackage{table}{xcolor}
\usepackage{colortbl}
\usepackage{booktabs}
\usepackage{multirow}
\usepackage[percent]{overpic} 
\usepackage{rotating}
\usepackage{subcaption}
\usepackage{wrapfig}
\usepackage{amssymb}

\usepackage{hyperref}

\usepackage{orcidlink}

\begin{document}

\title{SSR-GS: Separating Specular Reflection in Gaussian Splatting for Glossy Surface Reconstruction}

\titlerunning{SSR-GS}

\author{Ningjing Fan\inst{1} \and
Yiqun Wang\inst{1*}}

\institute{Chongqing University, Choingqing, China}

\maketitle

\begin{abstract}
  In recent years, 3D Gaussian splatting (3DGS) has achieved remarkable progress in novel view synthesis. However, accurately reconstructing glossy surfaces under complex illumination remains challenging, particularly in scenes with strong specular reflections and multi-surface interreflections. To address this issue, we propose SSR-GS, a specular reflection modeling framework for glossy surface reconstruction. Specifically, we introduce a prefiltered Mip-Cubemap to model direct specular reflections efficiently, and propose an IndiASG module to capture indirect specular reflections. 
  Furthermore, we design Visual Geometry Priors (VGP) that couple a reflection-aware visual prior via a reflection score (RS) to downweight the photometric loss contribution of reflection-dominated regions, with geometry priors derived from VGGT, including progressively decayed depth supervision and transformed normal constraints. Extensive experiments on both synthetic and real-world datasets demonstrate that SSR-GS achieves state-of-the-art performance in glossy surface reconstruction.
  \keywords{Gaussian Splatting \and Specular Reflection \and Surface Reconstruction}
\end{abstract}

\section{Introduction}
\label{sec:intro}

Accurate surface reconstruction from multi-view images remains a long-standing challenge in computer vision and graphics, with applications spanning animation, robotics, AR/VR, and autonomous driving. Neural Radiance Fields (NeRF)~\cite{Mildenhall2021} have recently achieved impressive fidelity~\cite{Liu2023, Ge2023, Wang2023}, but their dense neural representation incurs high computational costs and long training times. To address these limitations, 3D Gaussian Splatting (3DGS)~\cite{kerbl2023} models scenes using explicit 3D Gaussian primitives, enabling real-time rendering with high-quality view synthesis, thus offering a practical alternative for diverse applications.

\begin{figure*}[t]
\centering
\begin{overpic}[width=\linewidth]{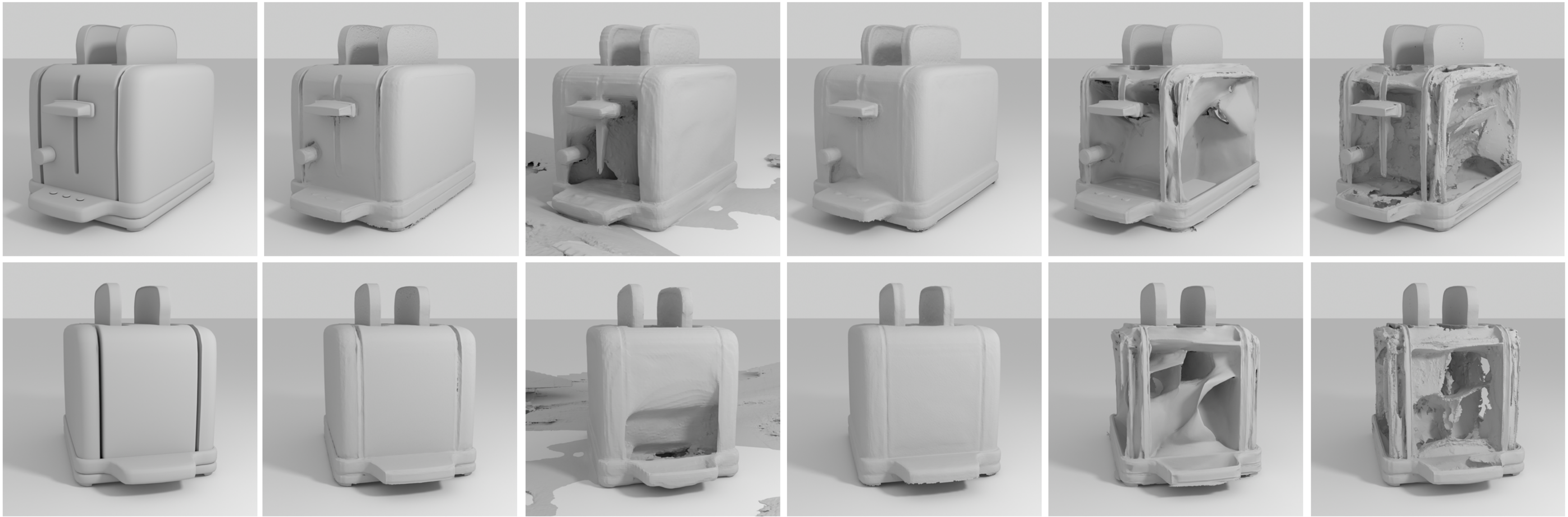}
    \put(6,-2.5){\fontsize{8pt}{0}\selectfont\color{black}{GT}}
    \put(22.5,-2.5){\fontsize{8pt}{0}\selectfont\color{black}{Ours}}
    \put(35.5,-2.5){\fontsize{8pt}{0}\selectfont\color{black}{Ref-Gaussian}}
    \put(54.6,-2.5){\fontsize{8pt}{0}\selectfont\color{black}{Ref-GS}}
    \put(71.8,-2.5){\fontsize{8pt}{0}\selectfont\color{black}{PGSR}}
    \put(89,-2.5){\fontsize{8pt}{0}\selectfont\color{black}{GOF}}

    \put(6,-5){\fontsize{8pt}{0}\selectfont\color{black}{CD}}
    \put(22.4,-5){\fontsize{8pt}{0}\selectfont\color{black}{0.067}}
    \put(39.4, -5){\fontsize{8pt}{0}\selectfont\color{black}{0.260}}
    \put(56,-5){\fontsize{8pt}{0}\selectfont\color{black}{0.072}}
    \put(72.3,-5){\fontsize{8pt}{0}\selectfont\color{black}{0.124}}
    \put(89,-5){\fontsize{8pt}{0}\selectfont\color{black}{0.109}}
\end{overpic}
\vspace{0cm}
\caption{Surface reconstruction results on toaster. CD denotes the Chamfer distance.}
\vspace{-0.4cm}
\label{fig:mesh}
\end{figure*}
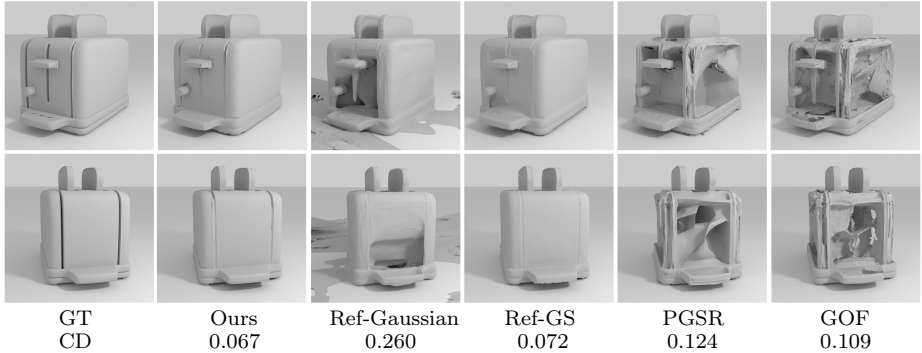

Although 3DGS enables real-time, high-quality novel view synthesis, it is primarily rendering-oriented and suffers from limited surface reconstruction accuracy due to the unstructured Gaussians and reliance on image reconstruction loss, which can lead to geometric deviation. 
Recent works have tried to improve geometric fidelity, such as SuGaR~\cite{Guedon2024}, 2DGS~\cite{Huang2024}, GOF~\cite{Yu2024GOF}, and PGSR~\cite{Chen2024}; and to enhance specular modeling, e.g., Spec-Gaussian~\cite{Yang2024}, 3DGS-DR~\cite{ye2024gsdr}, Ref-GS~\cite{Zhang2024}, and Ref-Gaussian~\cite{yao2024refGS}.
However, existing methods still struggle to faithfully reconstruct glossy surfaces with strong specular reflections (Fig.~\ref{fig:mesh}). In such cases, reflected radiance is often imperfectly separated from the diffuse component, which can cause light leakage and ultimately lead to geometric artifacts such as surface collapse in highly reflective regions.

In this work, we propose \textbf{SSR-GS}, a framework for separating specular reflection in Gaussian splatting for high-fidelity glossy surface reconstruction. 
We decouple diffuse and specular components: the diffuse term is integrated via volumetric compositing along the ray, while the specular term is factorized into a material component (via volumetric blending) and incident illumination estimated from physically based surface rendering.
Specular reflection is further decomposed into direct and indirect components: direct reflection is computed using a Mip-Cubemap with view-consistent environment sampling, and indirect reflection is modeled by the proposed IndiASG to capture complex multi-bounce effects while remaining separated from the diffuse term.
To enhance geometric fidelity and cross-view consistency, we incorporate Visual Geometry Priors (VGP), coupling a visual prior (VP) with geometry priors (GP). The VP, implemented via the reflection score (RS), suppresses reflection-dominated regions, while the GP applies VGGT-inferred depth and transformed normal constraints to regularize geometry, enabling more stable optimization and higher-quality reconstruction under complex reflections.

Our main contributions are:

\begin{itemize}
\item \textbf{Mip-Cubemap environment representation:} 
We propose a Mip-Cubemap environment representation for modeling direct specular reflections, enabling multi-scale environment sampling and more accurate roughness-aware reflection rendering.

\item \textbf{IndiASG for indirect specular reflection modeling:}
We identify that inadequate modeling of indirect specular reflections can destabilize geometry estimation. Based on this insight, we propose IndiASG, which explicitly models indirect specular reflections to improve geometric stability and enable Gaussians to better capture multi-view consistent geometry.

\item \textbf{Visual Geometry Priors (VGP):} We propose visual geometry priors that couple a visual prior reflection score (RS) with geometry priors inferred by VGGT, which work synergistically to constrain and stabilize geometry.
\end{itemize}

\section{Related Works}
\label{sec:related_works}

\subsection{Surface Reconstruction via Neural Rendering}
NeRF~\cite{Mildenhall2021} pioneered volumetric scene modeling in neural rendering, inspiring subsequent geometry-aware extensions such as signed distance fields (e.g., NeuS~\cite{Wang2021} and its variants~\cite{Fu2022,Ge2023,Wang_2023_CVPR, NEURIPS2022_0ce8e343}), occupancy-based formulations~\cite{Oechsle2021}, and reflective surface handling~\cite{Verbin2022,Liang2023,Liu2023,Fan2023}.

With the advent of 3D Gaussian Splatting (3DGS)~\cite{kerbl2023}, explicit 3D Gaussian primitives have been explored for surface reconstruction. However, due to the discrete and unstructured nature of Gaussians, accurate surface extraction remains challenging. Methods such as SuGaR~\cite{Guedon2024} introduce regularization terms to constrain Gaussians to surfaces, followed by Poisson reconstruction~\cite{Kazhdan2013}. NeuSG~\cite{Chen2023}, GSDF~\cite{Yu2024GSDF}, and 3DGSR~\cite{Lyu2024} optimize SDFs jointly with Gaussian models, but neural optimization is computationally expensive. 2DGS~\cite{Huang2024} employs 2D Gaussian primitives and TSDF fusion~\cite{Izadi2011} for surface extraction. GOF~\cite{Yu2024GOF} extracts geometry from Gaussian opacity fields without Poisson or TSDF fusion. PGSR~\cite{Chen2024} leverages unbiased depth rendering and integrates single- and multi-view geometry through regularization for accurate reconstruction. GausSurf~\cite{Wang2024} further incorporates multi-view constraints and normal priors~\cite{Ye2024} to improve quality while reducing computation.
MILo~\cite{guedon2025milo} introduces a differentiable Gaussian Splatting framework that extracts meshes directly during training, bridging volumetric and surface representations for accurate and lightweight surface reconstruction.
Despite these advances in surface extraction and geometric regularization, existing methods primarily focus on general scene reconstruction and do not explicitly address the challenges posed by glossy surfaces.

\subsection{Modeling Glossy Reflections}
GaussianShader~\cite{Jiang2024} enhances 3DGS rendering of reflective surfaces through a simplified shading function with reliable normal estimation. 3iGS~\cite{tang20243igs} improves the view-dependent specular quality by factorizing a continuous illumination field and optimizing per-Gaussian BRDF features. Relightable 3D Gaussians~\cite{R3DG2023} enables physically based relighting by jointly optimizing per-point normals, BRDFs, and incident lighting through differentiable rendering.
Spec-Gaussian~\cite{Yang2024} replaces SH-based colors with an anisotropic spherical Gaussian (ASG) appearance field, enabling improved modeling of high-frequency and anisotropic view-dependent specular effects.
EnvGS~\cite{xie2025envgs} models complex view-dependent reflections by introducing environment Gaussian primitives jointly optimized with scene Gaussians, rendered efficiently via GPU ray tracing.
Ref-GS~\cite{Zhang2024} integrates deferred rendering and directional encoding, reducing view-dependent ambiguities and introducing a spherical Mip-Grid to capture surface roughness. Ref-Gaussian~\cite{yao2024refGS} enables real-time reconstruction of reflective objects with inter-reflection via physically based deferred rendering and Gaussian-grounded inter-reflection modeling. 
IRGS~\cite{gu2024IRGS} models inter-reflections in inverse rendering via differentiable 2D Gaussian ray tracing and Monte Carlo optimization for indirect lighting estimation. GlossyGS~\cite{glossygs} and MaterialRefGS~\cite{zhang2025materialrefgs} address geometry–material ambiguity in reflective scenes through microfacet priors, multi-view consistent material inference, and environment modeling.  
GOGS~\cite{yang2025gogs} applies physics-based Gaussian surfels with geometric priors and refined specular modeling for high-fidelity glossy object reconstruction and relighting.  
Overall, these works progressively improve the modeling of view-dependent and reflective effects, ranging from enhanced shading functions to physically grounded inverse rendering and environment-aware illumination representations. 
However, accurately reconstructing geometry under strong specularities and complex lighting remains challenging, as reflected radiance can interfere with surface estimation and cause geometric artifacts. Our method mitigates this by improving the separation of specular reflections and stabilizing geometry optimization, preventing reflected structures from being baked into the reconstructed surface.

\section{Preliminary}
\label{sec:preliminary}

\subsection{3D Gaussian Splatting (3DGS)}
3D Gaussian splatting~\cite{kerbl2023} models a scene using a collection of 3D Gaussian primitives. Each Gaussian is characterized by a center $\boldsymbol{\mu} \in \mathbb{R}^{3}$, a covariance $\boldsymbol{\Sigma} \in \mathbb{R}^{3 \times 3}$, an opacity scalar $\alpha$, and view-dependent color coefficients $\boldsymbol{c}$ (represented via spherical harmonics).
The spatial contribution of a Gaussian is defined as:
\begin{equation}
    \mathcal{G}(\boldsymbol{x}) = \exp\!\left[ -\frac{1}{2} (\boldsymbol{x} - \boldsymbol{\mu})^{T} \boldsymbol{\Sigma}^{-1} (\boldsymbol{x} - \boldsymbol{\mu}) \right].
\end{equation}
Rendering is performed via splatting and point-based $\alpha$-blending. For a pixel, the accumulated color $\boldsymbol{C}$ is computed by traversing $K$ depth-sorted Gaussians overlapping the ray:
\begin{equation}
\boldsymbol{C} = \sum_{k=1}^K \boldsymbol{c}_k \,\alpha_k\, \mathcal{G}'_k
\prod_{j=1}^{k-1} \big( 1 - \alpha_j\, \mathcal{G}'_j \big),
\label{eq:volume_rendering}
\end{equation}
where $\mathcal{G}'$ represents the projected 2D Gaussian on the image plane.

\subsection{Physically Based Rendering}
Physically based rendering (PBR) models light transport according to physical principles. The outgoing radiance is described by the rendering equation~\cite{kajiya1986rendering}:
\begin{equation}
L_o(\mathbf{x}, \omega_o) = L_e(\mathbf{x}, \omega_o) +
\int_{\Omega} f_r(\mathbf{x}, \omega_i, \omega_o)
L_i(\mathbf{x}, \omega_i) (\mathbf{n} \cdot \omega_i) \, d\omega_i .
\end{equation}
We adopt a Cook--Torrance microfacet BRDF~\cite{cook1981reflectance} with a Disney-style energy-conserving formulation~\cite{burley2012physically}, decomposing the BRDF into diffuse and specular components:
\begin{equation}
f_\mathrm{diff}(\mathbf{x},\omega_i, \omega_o)
=(1-F(\omega_i,\mathbf{h}))(1-m)\frac{c_{\mathrm{albedo}}}{\pi},
\label{eq:diff_brdf}
\end{equation}
\begin{equation}
f_\mathrm{spec}(\mathbf{x},\omega_i, \omega_o)
=\frac{D(r, \mathbf{n}, \mathbf{h})
F(\omega_i,\mathbf{h})
G(\omega_i,\omega_o,\mathbf{h})}
{4(\mathbf{n}\cdot\omega_i)(\mathbf{n}\cdot\omega_o)}.
\end{equation}
where
$F(\omega_i,\mathbf{h})$ is the Fresnel term, for which the Schlick approximation is employed: 
  \begin{equation}
    F(\omega_i,\mathbf{h})\approx F_0+(1-F_0)(1-\omega_i\cdot\mathbf{h})^5,
    \label{eq:frenel}
    \end{equation}
    where $F_0$ denotes the Fresnel reflectance at normal incidence;
$m$ is the metalness;
$c_{\mathrm{albedo}}$ is the diffuse albedo color;
$\mathbf{h}=\frac{\omega_i+\omega_o}{\|\omega_i+\omega_o\|}$ is the half-vector;
$r$ is the roughness;
$D(r, \mathbf{n}, \mathbf{h})$ is the normal distribution function;
$G(\omega_i,\omega_o,\mathbf{h})$ is the geometry attenuation term.

Specular reflection can be factorized into a material-dependent term $M_{\mathrm{spec}}$ and a lighting-dependent term $I_{\mathrm{spec}}$~\cite{karis2013real}:  
\begin{equation}
L_o^\mathrm{spec}(\mathbf{x}, \omega_o) 
\approx 
\underbrace{\int_\Omega \frac{F\,G}{4(\mathbf{n}\cdot \omega_o)} \, d\omega_i}_{M_{\mathrm{spec}}}
\cdot
\underbrace{\int_\Omega L_i(\omega_i)\,D(\omega_i) \, d\omega_i}_{I_{\mathrm{spec}}}.
\label{eq:specular}
\end{equation}

\section{Method}
\label{sec:method}

\begin{figure*}[t]
\centering
\includegraphics[width = \linewidth]{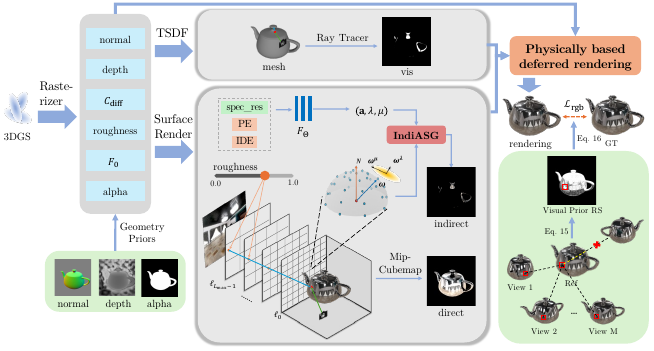}
\caption{An overview of our SSR-GS pipeline.
We rasterize 3D Gaussians to obtain surface buffers including normals, depth, opacity, diffuse component $C_{\mathrm{diff}}$, roughness, $F_0$, and alpha,  and supervise them with geometry priors (GP).
After rasterization, we extract a mesh via TSDF fusion and perform mesh-based ray tracing to estimate visibility $w_{\mathrm{vis}}$.
Direct specular reflection is queried from a Mip-Cubemap environment map, while indirect specular reflection is modeled by IndiASG.
Finally, we apply physically based deferred rendering (Eq.~\ref{eq:final_color}) to produce the rendered image.
We additionally compute the visual prior (VP) (reflection score) and use it to down-weight reflection-dominated pixels in the photometric loss for stable geometry initialization.}


\label{fig:pipeline}
\end{figure*}

Our framework builds upon a physically based rendering formulation with explicit specular modeling. 
We first present the overall framework in Sec.~\ref{sec:overview}. 
Direct and indirect specular reflections are modeled using a Mip-Cubemap (Sec.~\ref{sec:direct}) and IndiASG (Sec.~\ref{sec:indirect}), respectively. 
To enhance geometric stability in reflective regions, we incorporate Visual Geometry Priors (VGP) (Sec.~\ref{sec:VGP}). 
The overall pipeline is shown in Fig.~\ref{fig:pipeline}.

\subsection{Overview}
\label{sec:overview}

Each Gaussian is endowed with physically meaningful material parameters: the Fresnel reflectance at normal incidence $f_0\!\in\!\mathbb{R}^3$ and a diffuse component $c_{\text{diff}}\!\in\!\mathbb{R}^3$. 
Through volumetric rendering (Eq.~\ref{eq:volume_rendering}), we obtain pixel-wise $F_0$ and $C_{\text{diff}}$, where $C_{\text{diff}}$ corresponds to the diffuse BRDF term $(1-m)\frac{c_{\mathrm{albedo}}}{\pi}$ in Eq.~\ref{eq:diff_brdf}. 
The Fresnel term $F$ is computed via Eq.~\ref{eq:frenel}, and the final radiance $L_{rgb}$ is given by:
\begin{equation}
\begin{split}
    L_{rgb} &= L_{diff}+L_{spec}, \\
    L_{diff} &= (1-F)\, C_{\text{diff}}, \quad
    L_{spec} = M_{\mathrm{spec}} \left( (1-w_{\mathrm{vis}})\,L_{\mathrm{spec}}^{\mathrm{direct}} 
    + w_{\mathrm{vis}}\,L_{\mathrm{spec}}^{\mathrm{indi}} \right),
\end{split}
\label{eq:final_color}
\end{equation}
where $L_{\mathrm{spec}}^{\mathrm{direct}}$ and $L_{\mathrm{spec}}^{\mathrm{indi}}$ denote the direct and indirect specular radiance terms described in Sec.~\ref{sec:direct} and Sec.~\ref{sec:indirect}, respectively.
$w_{\mathrm{vis}}$ denotes the visibility weight obtained via ray tracing on the mesh reconstructed, which blends direct and indirect specular contributions. $M_{\mathrm{spec}}$ is a specular material-dependent term computed via Eq.~\ref{eq:specular}.


\subsection{Mip-Cubemap for Direct Specular Reflection}
\label{sec:direct}
We approximate direct specular reflection using a roughness-aware environment map query, avoiding the costly hemispherical integration in the rendering equation.
Specifically, we introduce a physically motivated normal distribution function (NDF) prefilter approximation, which reduces the original integral to a single Mip-Cubemap lookup.
We query the direct specular reflection from the environment map using a roughness-aware mipmap hierarchy 
$\{\ell_0, \dots, \ell_{L_{\max}-1}\}$, where $L_{\max}$ denotes the total number of mip levels in the environment map. 
Higher mip levels correspond to progressively blurrier representations, modeling broader specular lobes induced by surface roughness. 
The mip level $\ell$ is determined by the surface roughness $r$ as
\begin{equation}
    \ell = r^2 \cdot (L_{\max} - 1).
\label{eq:mip_level}
\end{equation}
Direct specular reflection is then obtained via trilinear sampling of the environment map over direction and mip level:
\begin{equation}
    L_{\mathrm{spec}}^{\mathrm{direct}} = E^{\mathrm{trilinear}}(\omega_r, \ell)
\end{equation}
where $E$ denotes the environment map and $E^{\mathrm{trilinear}}(\omega_r, \ell)$ queries it at reflection direction $\omega_r$ and mip level $\ell$ (bilinear filtering within each level and linear interpolation across adjacent levels). Mipmap prefiltering implicitly accounts for roughness-dependent lobe shaping, avoiding explicit angular integration.

Our Mip-Cubemap differs from the cubemap in 3DGS-DR~\cite{ye2024gsdr} and the Sph-Mip in Ref-GS~\cite{Zhang2024} in both representation and filtering.
3DGS-DR uses a standard cubemap without roughness-aware mip selection, whereas we build a mipmap hierarchy and choose mip levels based on roughness (Eq.~\ref{eq:mip_level}) for prefiltered specular sampling.
Compared with the spherical parameterization in Ref-GS, our cubemap avoids projection distortion and naturally supports mip-based filtering.
Ablations in Sec.~\ref{sec:ablation} validate this design.

\subsection{IndiASG for Indirect Specular Reflection}
\label{sec:indirect}
To address the challenge of inaccurate geometry caused by multi-surface indirect illumination, we propose IndiASG, a compact, learning-based local light field representation for indirect specular reflection. IndiASG models the reflection using a fixed set of anisotropic spherical Gaussian lobes, with a neural predictor $F_{\Theta}$ estimating the per-lobe radiometric parameters from a surface point and a reflection direction. This design enables physically consistent indirect specular modeling while maintaining accurate surface reconstruction.

At each surface point, indirect specular reflection is modeled as the sum of $N_{\mathrm{lobe}}=33$ anisotropic spherical Gaussian lobes over the upper hemisphere, i.e., the indirect specular illumination is represented as a superposition of these lobes.
Formally, each lobe $j$ is defined by 
$(\mathbf{a}_j, \lambda_j, \mu_j, \boldsymbol{\omega}_j, \boldsymbol{\omega}_j^{\lambda}, \boldsymbol{\omega}_j^{\mu})$, 
where $\mathbf{a}_j \in \mathbb{R}^3$ denotes RGB amplitude,  
$\lambda_j$ and $\mu_j$ control Gaussian sharpness along $\boldsymbol{\omega}_j^{\lambda}$ and $\boldsymbol{\omega}_j^{\mu}$, respectively,  
and $(\boldsymbol{\omega}_j, \boldsymbol{\omega}_j^{\lambda}, \boldsymbol{\omega}_j^{\mu})$ are precomputed unit vectors forming an orthonormal frame in tangent space.
The set of $\boldsymbol{\omega}_j$ follows a fixed hierarchical layout: one lobe at the zenith $\boldsymbol{\omega}_0 = (0,0,1),$ and four concentric rings at polar angles $\theta_\ell = \frac{\ell\,\pi}{8}, \quad \ell = 1,\dots,4,$
each containing eight evenly spaced azimuthal directions  
$\phi_k = \frac{2\pi k}{8}, \quad k = 0,\dots,7$ (Fig.~\ref{fig:pipeline}), with $(\boldsymbol{\omega}_j^{\lambda}, \boldsymbol{\omega}_j^{\mu})$ obtained via orthonormal completion of $\boldsymbol{\omega}_j$ in tangent space.

Given the predicted radiometric parameters $(\mathbf{a}_j, \lambda_j, \mu_j)$ and the precomputed geometric frame $(\boldsymbol{\omega}_j, \boldsymbol{\omega}_j^{\lambda}, \boldsymbol{\omega}_j^{\mu})$, the indirect specular reflection in a reflection direction ${\omega}_r$ is:
\begin{equation}
\begin{split}
L_{\mathrm{spec}}^{\mathrm{indi}}(\omega_r)
= \sum_{j=1}^{N_{\mathrm{lobe}}} \mathbf{a}_j\,
\max\!\left(0,\, \omega_r \cdot \boldsymbol{\omega}_j\right)
\exp\!\left(
-\lambda_j\, (\omega_r \cdot \boldsymbol{\omega}_j^{\lambda})^2
-\mu_j\, (\omega_r \cdot \boldsymbol{\omega}_j^{\mu})^2
\right),
\end{split}
\label{eq:asg}
\end{equation}
where $\max(0,\, {\omega}_r \cdot \boldsymbol{\omega}_j)$ 
enforces a hemisphere constraint. 
The exponential term models anisotropic angular falloff in the local tangent frame defined by 
$\boldsymbol{\omega}_j^{\lambda}$ and $\boldsymbol{\omega}_j^{\mu}$, 
with $\lambda_j$ and $\mu_j$ controlling the sharpness along the two principal axes.

We predict the per-lobe parameters via a learned mapping
\begin{equation}
F_{\Theta}: (\mathbf{p}, \boldsymbol{\omega}_r, r, C_{res})
\;\mapsto\; (\mathbf{a}_j, \lambda_j, \mu_j),
\end{equation}
where $\mathbf{p}$ and $\boldsymbol{\omega}_r$ are encoded with multi-frequency positional encoding (PE) and integrated directional encoding (IDE)~\cite{Verbin2022}, respectively; $r$ denotes the surface roughness; and $C_{res}$ denotes the residual specular signal.
The indirect specular radiance $L_{\mathrm{spec}}^{\mathrm{indi}}$ is then evaluated via Eq.~(\ref{eq:asg}).

\subsection{Visual Geometry Priors (VGP)}
\label{sec:VGP}
\subsubsection{Visual Prior (VP): Reflection Score (RS)}
\label{sec:reflection_score}

We take inspiration from Ref-NeuS~\cite{Ge2023} by employing a reflection score (RS) to identify regions with high multi-view variance. By attenuating the photometric optimization gradients in specular reflection regions, we reduce the adverse influence of view-dependent appearance variations on geometry updates, thereby ensuring stable surface reconstruction.

\paragraph{Multi-view Visibility and Occlusion}
Given a reference image $I_{\mathrm{ref}}$ and its rendered depth map $D_{\mathrm{ref}}$, we back-project each pixel $\mathbf{u}$ to a 3D surface point $\mathbf{p}$ in world coordinates. 
Let $\mathbf{r}_d(\mathbf{u})$ denote the ray direction in camera coordinates. The inverse projection $\pi^{-1}$ is formulated as:
\begin{equation}
    \mathbf{p} = \pi^{-1}(\mathbf{u}, D_{\mathrm{ref}}(\mathbf{u})) = \mathbf{R}^\top \left( D_{\mathrm{ref}}(\mathbf{u}) \cdot \mathbf{r}_d(\mathbf{u}) - \mathbf{t} \right),
\end{equation}
where $\mathbf{R}$ and $\mathbf{t}$ denote the rotation and translation of the world-to-camera extrinsic matrix, respectively. 

For a source view $m \in \mathcal{M}$ with camera center $\mathbf{o}_m$, projection $\pi_m$, and depth map $D_m$, we define a visibility indicator $\mathbb{1}_{m}(\mathbf{p})$. The point $\mathbf{p}$ is visible in view $m$ if:
\begin{enumerate}
    \item 	\textbf{Field of View:} The projected coordinate $\mathbf{u}_m = \pi_m(\mathbf{p})$ lies within the image boundaries.
    \item \textbf{Depth-based Occlusion:} The point $\mathbf{p}$ should not be occluded in view $m$. We perform a depth agreement check by comparing the point-to-camera distance with the depth map at its projected pixel:
    \begin{equation}
        \big| \| \mathbf{p} - \mathbf{o}_m \|_2 - D_m(\mathbf{u}_m) \big| < \tau_{\mathrm{occ}},
    \end{equation}
    where $\tau_{\mathrm{occ}}$ is a depth tolerance threshold (set to $0.15$ in our experiments).

    \item 	\textbf{View Sufficiency:} The point $\mathbf{p}$  must be visible in at least $K$ source views to calculate a reliable statistic (we use $K=5$).
\end{enumerate}

\paragraph{Reflection Score Formulation}
We define the reflection score $\mathrm{RS}(\mathbf{u})$ as the mean absolute photometric deviation between the reference view and valid source views. Let $\mathcal{V}(\mathbf{u}) = \{ m \in \mathcal{M} \mid \mathbb{1}_{m}(\mathbf{p}) = 1 \}$ be the set of valid views:
\begin{equation}
    \mathrm{RS}(\mathbf{u}) = \frac{1}{|\mathcal{V}(\mathbf{u})|} \sum_{m \in \mathcal{V}(\mathbf{u})} \big\| I_m(\mathbf{u}_m) - I_{\mathrm{ref}}(\mathbf{u}) \big\|_1.
\label{eq:rs}
\end{equation}
A higher reflection score indicates stronger view-dependent appearance changes, which are often caused by specular reflections and other non-Lambertian effects.

We use $I_{\mathrm{ref}}(\mathbf{u})$ as the anchor (instead of the mean over visible views) since $\mathrm{RS}$ later modulates the reference-view photometric loss (Eq.~\ref{eq:rs_loss}). This aligns $\mathrm{RS}$ with the reference view and suppresses the influence of view-dependent specularities on geometry reconstruction.

\paragraph{Reflection-aware Photometric Loss}
To reduce the adverse impact of specular reflections during early optimization, we incorporate RS into the training objective in Stage~1. 
The RS-weighted photometric loss is defined as:
\begin{equation}
    \mathcal{L}_{\mathrm{rgb}} = \frac{1}{|\Omega|} \sum_{\mathbf{u} \in \Omega} 
    \frac{\| \hat{C}(\mathbf{u}) - C_{\mathrm{gt}}(\mathbf{u}) \|_1}{\mathrm{RS}(\mathbf{u})},
\label{eq:rs_loss}
\end{equation}
where $\hat{C}$ and $C_{\mathrm{gt}}$ denote the rendered and ground-truth colors. 
Pixels with large reflection scores (i.e., high multi-view variance) are down-weighted, reducing the influence of view-dependent specular reflections on geometry updates. 

\subsubsection{Geometry Priors (GP)}

From VGGT~\cite{wang2025vggt}, we obtain a prior depth map $D_{\mathrm{VGGT}}$ and an associated confidence map $C_{\mathrm{VGGT}}$. 
We define a VGGT prior regularization that combines a depth consistency loss and a normal consistency loss, with per-pixel weights derived from $C_{\mathrm{VGGT}}$.

The depth term aligns the predicted depth map $D$ to the VGGT prior $D_{\mathrm{VGGT}}$:
\begin{equation}
    \mathcal{L}_{\mathrm{VGGT-D}} = \min_{\omega,\,b} \left\lVert \omega D + b - D_{\mathrm{VGGT}} \right\rVert_2^2.
\label{eq:vggt_depth}
\end{equation}
where the scalars $\omega$ and $b$ are estimated by least squares because the VGGT depth prior is not an absolute metric depth and differs by an unknown global scale and shift.

$N_{\mathrm{VGGT}}$ are estimated from $D_{\mathrm{VGGT}}$ by back-projecting pixels into 3D and computing the normalized cross product of vertical and horizontal offsets:
\begin{equation}
    N_{\mathrm{VGGT}} =
    \frac{\big(P_u - P_d\big) \times \big(P_r - P_l\big)}
    {\left\lVert \big(P_u - P_d\big) \times \big(P_r - P_l\big) \right\rVert_2},
\label{eq:depth2normal}
\end{equation}
where $P_u, P_d, P_l, P_r$ are the back-projected 3D point maps corresponding to neighboring pixels (up, down, left, and right), respectively.

The normal term combines an $\ell_1$ loss and an angular consistency loss between the predicted normal map $N$ and the VGGT-derived normal map $N_{\mathrm{VGGT}}$:
\begin{equation}
    \mathcal{L}_{\mathrm{VGGT-N}} =
    \left\lVert N - N_{\mathrm{VGGT}} \right\rVert_1
    + \lambda \left( 1 - \frac{N \cdot N_{\mathrm{VGGT}}}
    {\lVert N \rVert_2 \, \lVert N_{\mathrm{VGGT}} \rVert_2} \right),
\label{eq:vggt_normal}
\end{equation}
where $\lambda$ is the weighting factor of the angular consistency loss, and is set to $0.5$ in our experiments.

Per-pixel weights are obtained via a logarithmic transform followed by min--max normalization:
\begin{equation}
    W_{\mathrm{VGGT}} = \mathrm{Norm}\!\left( \log\!\left( 1 + C_{\mathrm{VGGT}} \right) \right).
\label{eq:vggt_weight}
\end{equation}

The final VGGT prior loss is defined as a confidence-weighted combination of the depth and normal terms:
\begin{equation}
    \mathcal{L}_{\mathrm{VGGT}} = W_{\mathrm{VGGT}} \cdot \left(
    \mathcal{L}_{\mathrm{VGGT-D}} + \mathcal{L}_{\mathrm{VGGT-N}}
    \right).
\label{eq:vggt_total}
\end{equation}

\begin{table}[t]
\centering
\scriptsize
\setlength{\tabcolsep}{4pt}
\renewcommand{\arraystretch}{1.2}
\definecolor{firstred}{HTML}{F4C7C3}
\definecolor{secondyellow}{HTML}{FFF2CC}
\begin{tabular}{l|ccccccc}
\hline
 & ball & car & coffee & helmet &
 teapot & toaster & \textbf{mean}\\
\hline
Ref-NeRF~\cite{Verbin2022}  & 1.55 & 14.93 & 12.24 & 29.48 & 9.23 & 42.87 & 18.38 \\
ENVIDR~\cite{Liang2023}        & 0.74 & 7.10 & 9.23 & 1.66  & 2.47 & 6.45 & 4.61 \\
UniSDF~\cite{Wang2023}        & \cellcolor{firstred} 0.45 & 6.88 & 8.00 & 1.72  & 2.80 & 6.45 & 8.71 \\
PGSR~\cite{Chen2024}& 66.93 & 4.62 & 2.91 & 6.01  & 1.01 & 15.31 & 16.13 \\
MILo~\cite{guedon2025milo}& 61.30 & 4.47 & 2.36 & 4.87  & 0.62 & 16.49 & 15.02 \\
GaussianShader~\cite{Jiang2024}& 7.03 & 14.05 & 14.93 & 9.33  & 7.17 & 13.08 & 10.93 \\
3DGS-DR~\cite{ye2024gsdr}      & 0.85 & 2.32 & 2.21 & 1.67 &  0.53 & 6.99 & 2.43 \\
Ref-Gaussian~\cite{yao2024refGS} &  0.71 & 1.91 & 2.34 & 1.85  & 0.48 & 5.70 & \cellcolor{secondyellow} 2.17 \\
Ref-GS~\cite{Zhang2024}        & 1.05 & 2.02 & 3.61 & 1.99 & 0.69 & \cellcolor{secondyellow} 3.92 & 2.21 \\
GS-ROR$^2$~\cite{Zhu2024} & \cellcolor{secondyellow} 0.47 & 2.06 & 5.47 & 1.82  & 0.52 & 5.52 & 2.64 \\
RTR-GS~\cite{zhou2025rtr} & 22.44 & \cellcolor{firstred}1.86 & 2.87 & \cellcolor{secondyellow}1.44  & 0.61 & 4.49 & 5.62 \\
MaterialRefGS~\cite{zhang2025materialrefgs}        & 0.63 & \cellcolor{firstred} 1.86 & \cellcolor{firstred} 1.79 & 1.46 & \cellcolor{secondyellow} 0.44 & 7.02 & 2.20 \\
Ours   & 0.77 & \cellcolor{secondyellow} 1.88 & \cellcolor{secondyellow} 1.97 & \cellcolor{firstred} 1.40 &  \cellcolor{firstred}0.43 & \cellcolor{firstred} 2.65 & \cellcolor{firstred} 1.52 \\ 
\hline
\end{tabular}
\caption{Quantitative surface reconstruction comparison on the ShinySynthetic dataset (normal MAE). CD is omitted since invisible inner surfaces in the GT meshes make it unreliable, as discussed in Ref-NeuS~\cite{Ge2023} for this dataset. Best results are highlighted in \cellcolor{firstred}{red} and second-best in \cellcolor{secondyellow}{yellow}.}
\label{tab:shiny_geo}
\vspace{-0.7cm}
\end{table}

\begin{table}[t]
\centering
\scriptsize
\setlength{\tabcolsep}{4pt}
\renewcommand{\arraystretch}{1.2}
\definecolor{firstred}{HTML}{F4C7C3}
\definecolor{secondyellow}{HTML}{FFF2CC}
\begin{tabular}{l|ccccccccc}
\hline
 & angel & bell & cat & horse &
 luyu & potion & tbell & teapot & \textbf{mean} \\
\hline
\multicolumn{10}{c}{CD$\downarrow$} \\
\hline
PGSR~\cite{Chen2024}& 0.77 & 3.08 & 3.39 & 0.90  & 1.16 & 3.99 & 4.59 & 4.57 & 2.81 \\
MILo~\cite{guedon2025milo}& 0.77 & 16.52 & 2.84 & 1.02  & 44.64 & 3.39 & 5.08 & 4.40 & 9.83 \\
GaussianShader~\cite{Jiang2024}& 0.85 & 1.10 & 2.56 & 0.73  & 1.07 & 4.74 & 5.74 & 3.40 & 2.53 \\
Ref-Gaussian~\cite{yao2024refGS} & 0.45 & 0.70 & 1.68 & 0.64  & \cellcolor{secondyellow} 0.88 & 0.81 & 0.59 & 1.01 & 0.95  \\
Ref-GS~\cite{Zhang2024}        & \cellcolor{secondyellow} 0.41 & 0.74 & 1.73 & 0.47 & 0.89 & 1.05 & \cellcolor{firstred} 0.52 & 0.88 & 0.84\\
GS-ROR$^2$~\cite{Zhu2024}        & 0.52 & \cellcolor{firstred} 0.32 & \cellcolor{secondyellow} 1.66 & \cellcolor{firstred} 0.46  & 0.92 & 0.86 & 0.58 & \cellcolor{firstred} 0.66 & \cellcolor{secondyellow} 0.75 \\
RTR-GS~\cite{zhou2025rtr} & 0.66 & 2.75 & 2.99 & 0.94  & 1.21 & 4.93 & 2.81 & 2.84 & 2.39  \\
MaterialRefGS~\cite{zhang2025materialrefgs} & 0.53 & 0.70 & 1.97 & 0.47 & 0.96 & \cellcolor{firstred} 0.62 & \cellcolor{secondyellow} 0.55 & 1.02 & 0.85\\
Ours   & \cellcolor{firstred} 0.35 & \cellcolor{secondyellow} 0.64 & \cellcolor{firstred} 0.59 & \cellcolor{secondyellow} 0.40 & \cellcolor{firstred} 0.70 & \cellcolor{secondyellow} 0.71 & 0.66 & \cellcolor{secondyellow} 0.76 & \cellcolor{firstred} 0.60  \\ 
\hline
\multicolumn{10}{c}{MAE$\downarrow$} \\
\hline
PGSR~\cite{Chen2024}& 4.90 & 6.39 & 8.43 & 5.97  & 5.91 & 11.12 & 12.62 & 6.28 & 7.70 \\
MILo~\cite{guedon2025milo}& 4.61 & 11.52 & 5.55 & 6.52  & 5.97 & 10.45 & 13.16 & 7.09 & 8.11 \\
GaussianShader~\cite{Jiang2024}& 2.90 & 1.60 & 4.33 & 3.27  & 4.56 & 9.52 & 5.60 & 3.01 & 4.35 \\
3DGS-DR~\cite{ye2024gsdr}       & 4.46 & 4.53 & 4.64 & 6.59 & 5.65 & 4.44 & 5.39 & 3.38 & 4.89\\
Ref-Gaussian~\cite{yao2024refGS} & \cellcolor{firstred} 1.79 & 1.16 & 3.15  & 4.03 & 3.15 & 3.04 & 2.02 & 1.20 & 2.44  \\
Ref-GS~\cite{Zhang2024}        & \cellcolor{secondyellow} 1.99 & 0.92 & 2.93 & \cellcolor{secondyellow} 3.18 & \cellcolor{secondyellow} 2.82 & 3.64 & 1.87 & 1.18 & 2.32\\
GS-ROR$^2$~\cite{Zhu2024}        & 2.09 & \cellcolor{secondyellow} 0.86 & 3.37 & 2.85 & 3.15 & 4.04 & 2.64 & \cellcolor{secondyellow} 1.04 & 2.51 \\
RTR-GS~\cite{zhou2025rtr} & 3.01 & 3.05 & 2.98 & 6.53  & 3.67 & 7.95 & 4.21 & 2.09 & 4.19  \\
MaterialRefGS~\cite{zhang2025materialrefgs} & 1.81 & \cellcolor{secondyellow} 0.86 & \cellcolor{secondyellow} 2.42 & \cellcolor{firstred} 3.16 & 3.07 & \cellcolor{secondyellow} 2.68 & \cellcolor{firstred} 1.77 & 1.18 & \cellcolor{secondyellow} 2.12\\
Ours   & 2.15 & \cellcolor{firstred} 0.75 & \cellcolor{firstred} 1.50 & 3.59 & \cellcolor{firstred} 2.78 & \cellcolor{firstred} 2.84 & \cellcolor{secondyellow} 1.80 & \cellcolor{firstred} 1.00 & \cellcolor{firstred} 2.05 \\
\hline
\end{tabular}
\caption{Quantitative surface reconstruction comparison on the GlossySynthetic dataset (CD$\times 10^2$ and normal MAE). Best results are highlighted in \cellcolor{firstred}{red} and second-best in \cellcolor{secondyellow}{yellow}.}
\label{tab:glossy_geo}
\vspace{-0.6cm}
\end{table}

\section{Training}
\label{sec:training}
\setlength{\intextsep}{0pt}
\begin{wraptable}{r}{0.42\textwidth}
\centering
\scriptsize
\setlength{\tabcolsep}{2pt}
\renewcommand{\arraystretch}{1.2}
\begin{tabular}{l|ccc}
\hline
& Mip-Cubemap & IndiASG & VGP \\
\hline
Stage~1 & $\checkmark$ & & $\checkmark$ \\
Stage~2 & $\checkmark$ & $\checkmark$ &  \\ 
\hline
\end{tabular}
\vspace{-0.2cm}
\caption{Two-stage strategy.}
\label{tab:ablation}
\end{wraptable}
As shown in Tab.~\ref{tab:ablation}, we employ a two-stage optimization strategy to progressively optimize geometry and illumination.
In Stage~1, we perform geometry initialization with Visual Geometry Priors (VGP), and use RS to downweight the photometric loss in the reflection-dominated region. Indirect illumination is disabled, and the specular reflection term is simplified to direct specular reflection:
$L_{\mathrm{spec}} = M_{\mathrm{spec}} \cdot L_{\mathrm{spec}}^{\mathrm{direct}}$.
In Stage~2, we enable indirect illumination using the IndiASG model and adopt the full rendering described in Sec.~\ref{sec:overview}, in which the VGP reweighting is disabled to allow full photometric supervision.



\section{Experiment}
\label{sec:experiment}

\begin{figure*}[t]
\centering
\begin{overpic}[width=\linewidth]{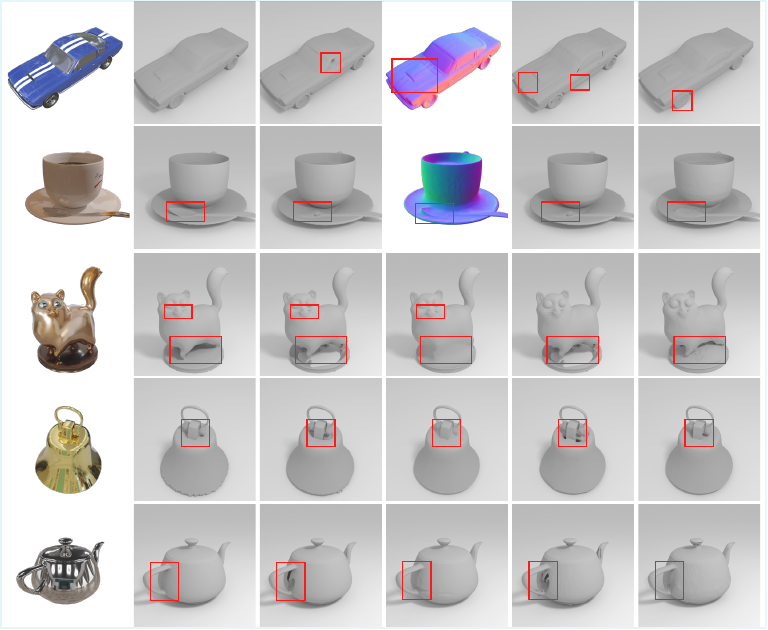}
    \put(5.5,-2){\fontsize{8pt}{0}\selectfont\color{black}{Image}}
    \put(22.5,-2){\fontsize{8pt}{0}\selectfont\color{black}{Ours}}
    \put(34,-2){\fontsize{8pt}{0}\selectfont\color{black}{MaterialRefGS}}
    \put(53.5,-2){\fontsize{8pt}{0}\selectfont\color{black}{GS-ROR$^2$}}
    \put(68,-2){\fontsize{8pt}{0}\selectfont\color{black}{Ref-Gaussian}}
    \put(87,-2){\fontsize{8pt}{0}\selectfont\color{black}{Ref-GS}}
    
\end{overpic}
\vspace{-0.2cm}
\caption{Qualitative results of surface reconstruction on ShinySynthetic (car, coffee) and GlossySynthetic (cat, bell, teapot) datasets. Since the released GS-ROR$^2$ code cannot extract meshes on the ShinySynthetic, we instead visualize its normal results for comparison.}

\label{fig:mesh1}
\end{figure*}

\begin{figure*}[t]
\centering
\begin{overpic}[width=\linewidth]{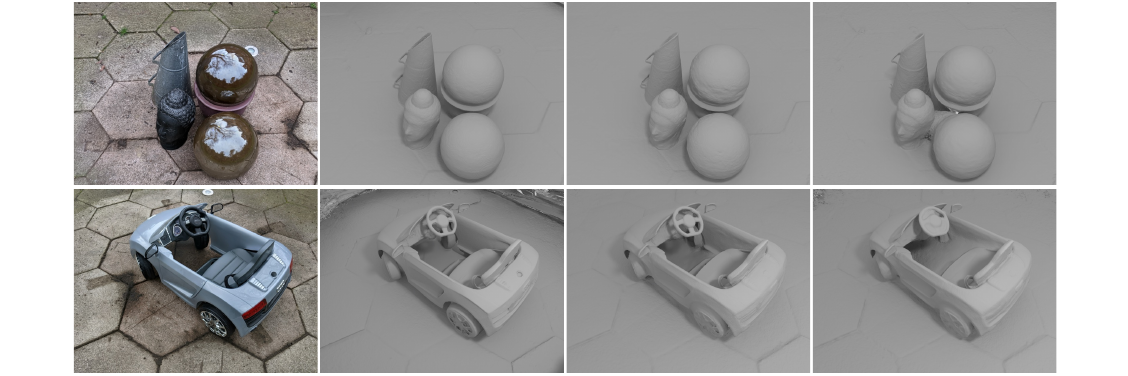}
    \put(13.5,-2){\fontsize{8pt}{0}\selectfont\color{black}{Image}}
    \put(36.5,-2){\fontsize{8pt}{0}\selectfont\color{black}{Ours}}
    \put(54.5,-2){\fontsize{8pt}{0}\selectfont\color{black}{Ref-Gaussian}}
    \put(79,-2){\fontsize{8pt}{0}\selectfont\color{black}{Ref-GS}}
    
\end{overpic}
\vspace{-0.3cm}
\caption{Qualitative results of surface reconstruction on Ref-Real (gardenspheres, toycar) dataset.}
\label{fig:mesh2}
\end{figure*}

\subsection{Implementation Details}
Our training process consists of two stages: Stage~1 runs for 10{,}000 iterations with densification at iteration 8{,}000, and Stage~2 runs for 20{,}000 iterations.  
The resolution of the cubemap is $6 \times 128 \times 128$ with four mipmap levels, downsampled by 4. 
$F_{\Theta}$ consists of three hidden layers with 128 units each. PE uses a frequency of 6, while IDE uses a frequency of 4.
All experiments are conducted on a single NVIDIA RTX~4090 GPU with 24\,GB of memory.

\subsection{Comparison}
\subsubsection{Datasets and Evaluation Metrics}

We evaluate our method on two synthetic datasets, ShinySynthetic~\cite{Verbin2022} and GlossySynthetic~\cite{Liu2023}, and a real-world dataset, Ref-Real~\cite{Verbin2022}.  
All scenes feature glossy surfaces with strong specular reflections and multi-surface indirect reflections, posing significant challenges for accurate surface reconstruction.
Geometry accuracy is evaluated using the mean angular error (MAE) of surface normals and the Chamfer distance (CD) of the reconstructed mesh.

\subsubsection{Comparisons}
We quantitatively evaluate performance on the ShinySynthetic dataset in Tab.~\ref{tab:shiny_geo} and the GlossySynthetic dataset in Tab.~\ref{tab:glossy_geo}. Since the Ref-Real dataset does not provide ground-truth meshes, quantitative geometric evaluation is not available. 
The results demonstrate that our surface reconstruction achieves state-of-the-art performance.
Qualitative evaluation on the ShinySynthetic and GlossySynthetic datasets is shown in Fig.~\ref{fig:mesh1}. In the \textbf{car}, our method avoids surface bumps in strongly textured regions, while correctly capturing the concave structure of the tires. In the \textbf{coffee}, the region between the spoon and the cup exhibits complex indirect illumination and shadowing; our model accurately reconstructs these challenging areas, yielding high-fidelity geometry. In the \textbf{cat}, our method cleanly separates the cat from its base without undesired connections and faithfully reconstructs fine structures such as the whiskers. In the \textbf{bell} and \textbf{teapot} scenes, the surfaces are highly glossy and exhibit strong intra-object indirect specular reflections. Our method reconstructs these regions with high-fidelity surfaces, avoiding erroneous geometric artifacts.


\subsection{Ablation Study}
\label{sec:ablation}








\begin{table}[t]
\centering
\scriptsize
\setlength{\tabcolsep}{0.5pt}
\renewcommand{\arraystretch}{1.2}

\begin{tabular}{l|c|c|c|c|c|c|c|c|c}
\hline

& \multicolumn{2}{c|}{Mip-Cubemap}
& \multicolumn{1}{c|}{IndiASG}
& \multicolumn{5}{c|}{VGP}
& Full \\
\cline{2-9}

& w/o Mip & w/o Cube
& w/o IASG
& w/o VGP & w/o VP & w/o GP & w/o GP-D & w/o GP-N
& Model \\
\hline

MAE$\downarrow$ & 1.56 & 1.62 & 1.52 & 2.25 & 1.55 & 2.15 & 1.53 & 2.86 & \textbf{1.52} \\
\hline
MAE$\downarrow$  & 2.07 & 2.17 & 2.06 & 2.78 & 2.07 & 2.74 & 2.05 & 2.72 & \textbf{2.05} \\
CD$\downarrow$ & 0.66 & 0.69 & 0.64 & 0.97 & 0.66 & 0.96 & 0.66 & 0.96 & \textbf{0.60} \\

\hline
\end{tabular}

\caption{Ablation study on different components. The first row corresponds to the ShinySynthetic dataset, while the remaining rows to the GlossySynthetic dataset.}
\vspace{-0.8cm}
\label{tab:ablation}
\end{table}

\begin{figure}[t]
\centering
\begin{overpic}[width=\linewidth]{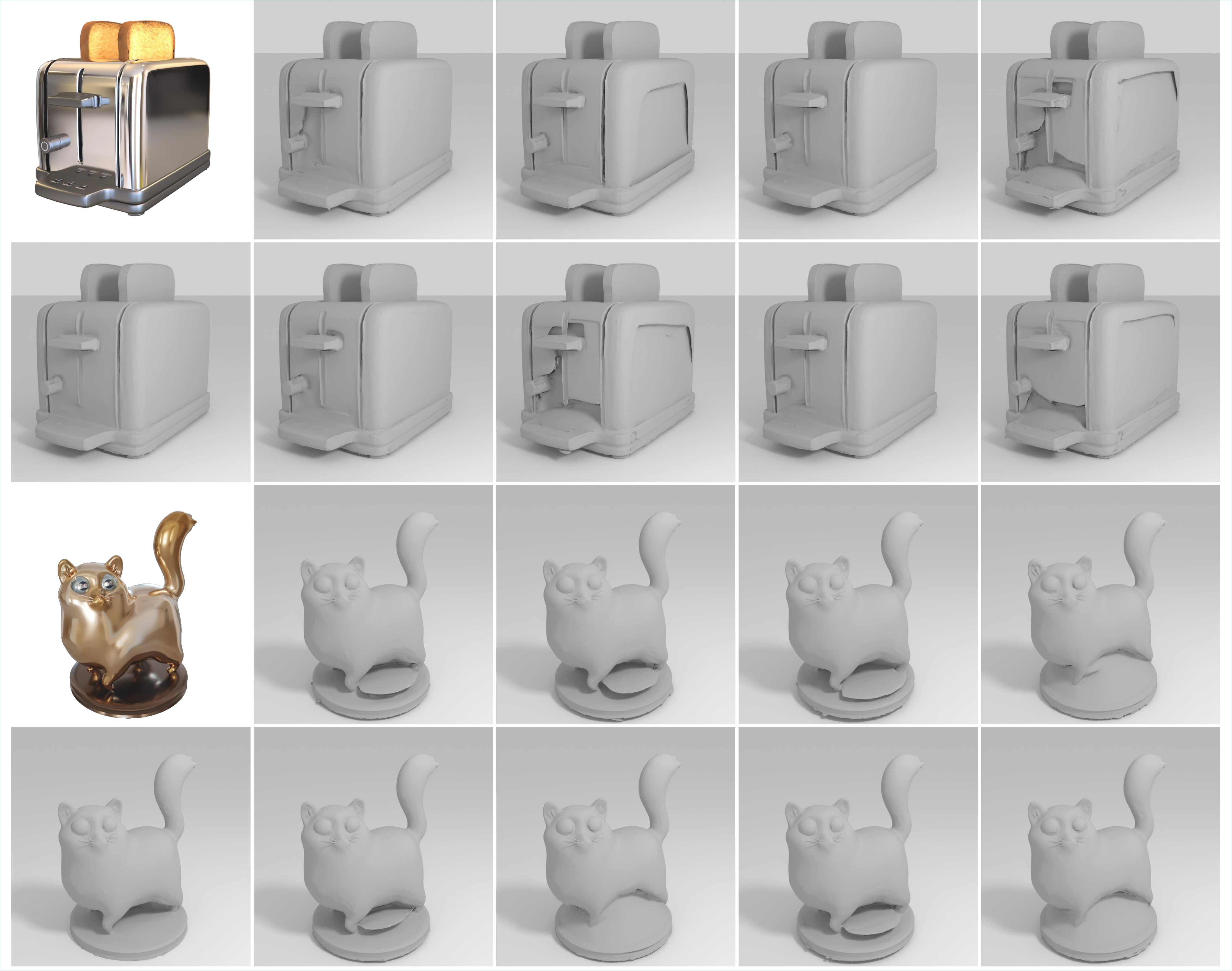}
    \put(0.5,37.8){\colorbox{white}{\fontsize{5pt}{0}\selectfont\color{black}{Image}}}
    \put(0.5,18){\colorbox{white}{\fontsize{5pt}{0}\selectfont\color{black}{Full Model}}}
    \put(20.5,37.8){\colorbox{white}{\fontsize{5pt}{0}\selectfont\color{black}{w/o Mip}}}
    \put(20.5,18){\colorbox{white}{\fontsize{5pt}{0}\selectfont\color{black}{w/o VP}}}
    \put(40.2,37.8){\colorbox{white}{\fontsize{5pt}{0}\selectfont\color{black}{w/o Cube}}}
    \put(40.2,18){\colorbox{white}{\fontsize{5pt}{0}\selectfont\color{black}{w/o GP}}}
    \put(59.7,37.8){\colorbox{white}{\fontsize{5pt}{0}\selectfont\color{black}{w/o IASG}}}
    \put(59.7,18){\colorbox{white}{\fontsize{5pt}{0}\selectfont\color{black}{w/o GP-D}}}
    \put(79.5,37.8){\colorbox{white}{\fontsize{5pt}{0}\selectfont\color{black}{w/o VGP}}}
    \put(79.5,18){\colorbox{white}{\fontsize{5pt}{0}\selectfont\color{black}{w/o GP-N}}}

    \put(0.5,77){\colorbox{white}{\fontsize{5pt}{0}\selectfont\color{black}{Image}}}
    \put(0.5,57.2){\colorbox{white}{\fontsize{5pt}{0}\selectfont\color{black}{Full Model}}}
    \put(20.5,77){\colorbox{white}{\fontsize{5pt}{0}\selectfont\color{black}{w/o Mip}}}
    \put(20.5,57.2){\colorbox{white}{\fontsize{5pt}{0}\selectfont\color{black}{w/o VP}}}
    \put(40.2,77){\colorbox{white}{\fontsize{5pt}{0}\selectfont\color{black}{w/o Cube}}}
    \put(40.2,57.2){\colorbox{white}{\fontsize{5pt}{0}\selectfont\color{black}{w/o GP}}}
    \put(59.7,77){\colorbox{white}{\fontsize{5pt}{0}\selectfont\color{black}{w/o IASG}}}
    \put(59.7,57.2){\colorbox{white}{\fontsize{5pt}{0}\selectfont\color{black}{w/o GP-D}}}
    \put(79.5,77){\colorbox{white}{\fontsize{5pt}{0}\selectfont\color{black}{w/o VGP}}}
    \put(79.5,57.2){\colorbox{white}{\fontsize{5pt}{0}\selectfont\color{black}{w/o GP-N}}}
\end{overpic}
\vspace{-0.5cm}
\caption{Ablation study on different components.}
\vspace{-0.5cm}
\label{fig:ablation}
\end{figure}


We perform ablation experiments on ShinySynthetic~\cite{Verbin2022} and GlossySynthetic~\cite{Liu2023} to evaluate the contribution of the individual components in our framework, thereby validating their effectiveness. Quantitative results are reported in Tab.~\ref{tab:ablation}, and qualitative comparisons are shown in Fig.~\ref{fig:ablation}.

\subsubsection{(1) Mip-Cubemap:}
We ablate our Mip-Cubemap environment representation with two variants: \textbf{w/o Mip}, which queries a single-resolution cubemap (akin to the Cubemap in 3DGS-DR~\cite{ye2024gsdr}), and \textbf{w/o Cube}, which replaces the cubemap with a spherical mipmap representation (akin to Sph-Mip in Ref-GS~\cite{Zhang2024}).
As shown in Tab.~\ref{tab:ablation} and Fig.~\ref{fig:ablation},
combining the cubemap with mip-level querying provides the best quality.

\subsubsection{(2) IndiASG:}
We introduce IndiASG to explicitly model indirect specular reflection (Sec.~\ref{sec:indirect}). 
To evaluate its effectiveness, we conduct an ablation study by removing the indirect illumination modeling (\textbf{w/o IASG}), relying solely on the Mip-Cubemap representation for specular rendering. 
As shown in Tab.~\ref{tab:ablation}, the full model shows a slight advantage over \textbf{w/o IASG}, though the improvement is marginal in the averaged metrics. This advantage can be understated because the scenes contain only small regions with indirect specular reflections, so the quantitative results are less sensitive to IndiASG's gains. Moreover, normal MAE may miss geometric errors; e.g., in the \textbf{cat} scene (Fig.~\ref{fig:ablation}), \textbf{w/o IASG} produces a misaligned base while the normals remain largely correct, yielding a small normal MAE despite noticeable geometry errors.

\subsubsection{(3) VGP:}
To comprehensively evaluate the role of Visual Geometry Priors (VGP), we consider five variants: \textbf{w/o VGP} removes both VP and GP; \textbf{w/o VP} removes VP while retaining GP; \textbf{w/o GP} removes GP while retaining VP; \textbf{w/o GP-D} removes the depth term $\mathcal{L}_{\mathrm{VGGT-D}}$ and keeps only the normal term $\mathcal{L}_{\mathrm{VGGT-N}}$ within GP; and \textbf{w/o GP-N} removes the normal term $\mathcal{L}_{\mathrm{VGGT-N}}$ and keeps only the depth term $\mathcal{L}_{\mathrm{VGGT-D}}$ within GP.

As shown in Fig.~\ref{fig:ablation}, removing both priors (\textbf{w/o VGP}) results in the most severe degradation, with unstable geometry and noticeable surface artifacts in reflective regions. When only the visual prior is removed (\textbf{w/o VP}), strong specular gradients interfere with early-stage optimization, leading to distorted structures in areas with multi-surface reflections (cat). Similarly, removing only the geometry prior (\textbf{w/o GP}) causes surfaces under complex illumination to become less consistent and less smooth, even though RS still mitigates part of the specular interference.
When using only a single geometric cue, performance drops compared to the full model. In \textbf{w/o GP-D} (normal only), the reconstructed geometry exhibits locally correct normals but suffers from depth layering artifacts, leading to stratified or offset surfaces along the viewing direction. In \textbf{w/o GP-N} (depth only), although the global depth structure is relatively preserved, local surface orientation becomes less accurate, and fine geometric details are weakened. In contrast, our full model jointly leverages both depth and normal supervision, achieving geometrically consistent surfaces with accurate global structure and sharp local detail. These results demonstrate that our proposed VGP combines VP and GP to yield complementary benefits and is essential for high-quality surface reconstruction under complex reflective conditions.

\section{Conclusion}
\label{sec:conclusion}

We propose SSR-GS, a framework for separating specular reflection in Gaussian Splatting for glossy surface reconstruction. 
Our method mitigates geometric artifacts in surface reconstruction under strong reflections and complex lighting.
Specifically, we explicitly decouple diffuse and specular components and further decompose specular reflection into direct and indirect terms. Direct reflection is modeled via the proposed Mip-Cubemap representation for roughness-aware, view-consistent environment map sampling, while indirect reflection is modeled using the proposed IndiASG, enabling accurate representation of complex multi-bounce illumination. In addition, we incorporate a hybrid Visual Geometry Priors (VGP), including a reflection-aware suppression mechanism based on the Reflection Score (RS) and complementary depth–normal constraints from VGGT, to improve geometric fidelity. Extensive experiments demonstrate that SSR-GS achieves state-of-the-art performance in geometry reconstruction.



%
%
\bibliographystyle{splncs04}
\bibliography{main}

\end{document}